\pdfoutput=1

\documentclass[11pt]{article}

\usepackage{acl}

\usepackage{times}
\usepackage{latexsym}

\usepackage[T1]{fontenc}

\usepackage[utf8]{inputenc}

\usepackage{microtype}

\usepackage{inconsolata}
\usepackage{multirow}
\usepackage{booktabs}

\usepackage{graphicx}
\usepackage{amsmath}
\usepackage{algorithm}
\usepackage{algpseudocode}
\usepackage{enumitem}
\usepackage{tcolorbox}
\usepackage{makecell}
\usepackage{pifont}
\tcbuselibrary{breakable}

\title{RoleRAG: Enhancing LLM Role-Playing via Graph Guided Retrieval}

\author{Yongjie Wang$^1$ 
  \and
  Jonathan Leung$^1$
  \and
  Zhiqi Shen$^2$ \\
  $^1$Alibaba-NTU Global e-Sustainability CorpLab (ANGEL),
  Nanyang Technological University\\
  $^2$College of Computing and Data Science,
  Nanyang Technological University\\
  \texttt{\{yongjie.wang, jonathan.leung, zqshen\}@ntu.edu.sg} \\}

\begin{document}
\maketitle
\begin{abstract} 
Large Language Models (LLMs) have shown promise in character imitation, enabling immersive and engaging conversations. However, LLMs often generate content that is irrelevant or inconsistent with a character’s background. We attribute these failures to: 1) the inability to accurately recall character-specific knowledge due to entity ambiguity; and 2) a lack of awareness of the character’s cognitive boundaries. This paper introduces RoleRAG, a retrieval-based framework that combines efficient entity disambiguation for knowledge indexing with a boundary-aware retriever to extract contextually appropriate content from a structured knowledge graph. We conducted extensive experiments on role-playing benchmarks and demonstrate that RoleRAG's calibrated retrieval enables both general LLMs and role-specific LLMs to exhibit knowledge that is more aligned with the given character and reduce hallucinated responses. 
 
\end{abstract}
\section{Introduction}

The advent of Large Language Models (LLMs) has significantly enhanced the capabilities of conversational AI agents due to their proficiency in understanding and generation. To further promote user engagement and entrainment \cite{park2023generative}, role-playing LLMs are designed to mimic the traits and experiences of specific characters, producing interactions that are role-consistent, emotionally deep, and contextually aware.  

\begin{figure}[t]
    \centering
     \includegraphics[width=\linewidth]{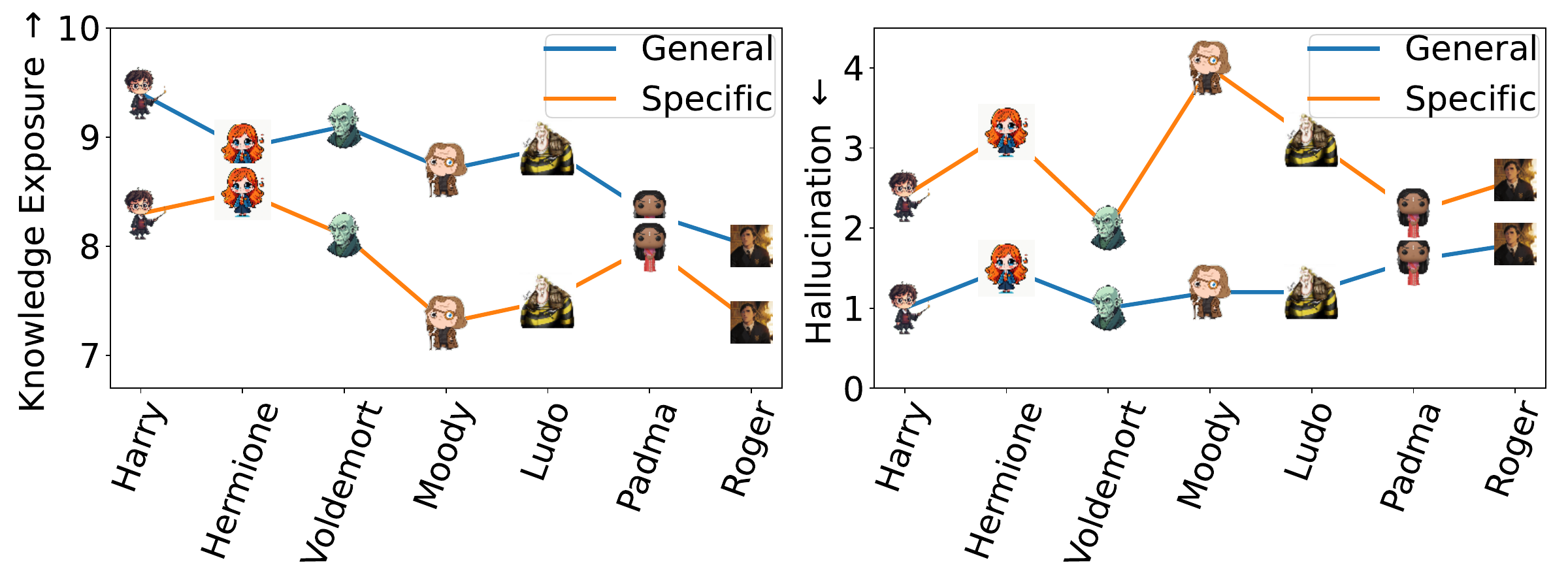}
     \caption{This figure illustrates that LLMs perform worse on role-specific questions, particularly when imitating lower-frequency characters.}
    \label{fig:hallucination}
    \vspace{-0.5cm}
\end{figure}

To improve imitation capabilities, recent studies~\cite{shao-etal-2023-character, tu-etal-2024-charactereval, tao-etal-2024-rolecraft, lu-etal-2024-large, zhou-etal-2024-characterglm} have fine-tuned LLMs on datasets specifically curated for role-playing scenarios. However, due to the labor-intensive nature of data collection and the high computational costs associated with fine-tuning, an alternative line of research explores the use of in-context learning by, for example, providing few-shot examples~\cite{li2023chatharuhi} or using static user profiles~\cite{wang-etal-2024-rolellm}, to provide role-related background information. 

However, LLMs still struggle to align accurately with a character's knowledge, often generating responses that lack appropriate traits or include fabricated content—particularly in certain role-playing scenarios where factual consistency is critical. As shown in Figure~\ref{fig:hallucination}, we tasked GPT-4o-mini with playing seven characters from the Harry Potter series, selected based on their frequency of appearance. Each character was presented with 10 general questions (e.g., interests, attitudes) and 10 role-specific questions (e.g., experiences, activities). We then recruited human raters to assess whether the language models accurately reflected each character's traits and to rate the severity of hallucinations, using a ten-point scale. From these results, we observe that LLMs perform worse on role-specific questions that require detailed character knowledge, particularly for less frequent characters. 

Our failure case analysis reveals two key factors contributing to these issues. (1) \textit{Entity ambiguity during knowledge extraction:} a single character may be referred to by different names across different stages or contexts. For example, `Anakin Skywalker' is also known as `Darth Vader' or `Lord Vader' in various installments of the Star Wars series. If not properly unified, such name variation can cause critical information to be missed during knowledge retrieval ; (2) \textit{Character-related cognition-boundary unawareness:} LLMs encode vast amounts of knowledge beyond the scope of the character they are portraying and often rely on LLM internal knowledge when responding to user queries. This can result in fabricated responses, particularly when the question falls outside the character’s original knowledge boundaries. Such hallucinations are unique to the role-playing setting.

To address these issues, we introduce RoleRAG, a retrieval-based framework specifically designed for role-playing tasks. Our approach is built on knowledge graph–enhanced retrieval, motivated by the observation that answering a single question may require reasoning over a broad range of dispersed textual knowledge. In the context of role-playing, the knowledge graph is constructed from character-centric corpora such as Wikipedia profiles and books. Each node in the graph represents an entity (e.g., character, location), and each edge encodes a semantic relationship between two entities (e.g., interactions between characters). To normalize duplicated names referring to the same entity, we propose an efficient semantic entity normalization algorithm. It first links name variants based on their local context, then clusters them into groups representing the same entity. Finally, an LLM is prompted to generate a unified canonical name for each group. The knowledge graph is then constructed using these normalized entities.

Our retrieval module, built on this graph-based indexing system, is designed to extract both specific and general entities mentioned in user queries while rejecting those that fall outside the scope of the character's knowledge. Subsequently, information relevant to the designated role is retrieved from the knowledge graph and provided to the LLM, equipping it with detailed contextual information to generate accurate responses. In contrast, out-of-scope questions are encouraged to be rejected to prevent the model from generating hallucinated or fabricated content.

Our contributions can be summarized as follows:
\begin{itemize}[noitemsep,topsep=0pt,leftmargin=*]
    \item We propose an efficient entity normalization algorithm that merges duplicated names referring to the same entity, thereby facilitating high-quality graph-based indexing over large character corpora.
    \item We introduce an effective retrieval module that not only retrieves both general concepts and character-specific details, but also helps reject out-of-scope questions to reduce hallucinations. 
    \item Extensive experiments that demonstrate that RoleRAG outperforms relevant baselines by exhibiting aligned character knowledge and reducing hallucinations.
\end{itemize}



\section{Related Works}
\textbf{LLM-based Role-Play} enables LLMs to embody user-specified characters, enhancing engagement through conversation. Existing research falls into three main directions: 
(1) Fine-tuning-based approaches \cite{shao-etal-2023-character,tu-etal-2024-charactereval,wang-etal-2024-rolellm,zhou-etal-2024-characterglm,lu-etal-2024-large} involve training open-source LLMs on curated character corpora. The training data is either synthetic—generated specifically for character conditioning \cite{shao-etal-2023-character,tu-etal-2024-charactereval,wang-etal-2024-rolellm}—or extracted from real-world datasets using LLMs \cite{zhou-etal-2024-characterglm}. (2) Retrieval-based approaches \cite{salemi-etal-2024-lamp,weir-etal-2024-ontologically,zhou-etal-2024-characterglm} fetch relevant documents from a character corpus to serve as contextual input to the LLM, thereby enhancing its ability to generate accurate and character-specific responses. The performance of these methods heavily depends on the quality and relevance of the retrieved content. (3) Plugin-based methods \cite{Liu2024} freeze the LLM while encoding each user's characteristics using a lightweight plugin model. The resulting user embedding is then concatenated with the embedding of the user's query to guide the LLM in generating personalized responses. A comprehensive comparison of the three categories is provided in the Appendix \ref{sec:appendix_a}. In this work, we follow retrieval-based approaches, aiming to provide character-relevant content while reducing hallucinations in responses. 

\textbf{Persona-based Dialogue.} Persona-based dialogue tasks require LLMs to exhibit general human-like traits such as humor, empathy, or curiosity, rather than adhering to specific role characteristics. Unlike role-playing, the focus is on consistent personality expression. Personas can be assigned via Big Five trait prompts \cite{jiang2023personallm}, character profiles \cite{tu-etal-2024-charactereval,zhou-etal-2024-characterglm}, or dialogue history \cite{zhong-etal-2022-less}. Evaluation is typically conducted through personality assessments or interviews \cite{wang-etal-2024-incharacter}. A comprehensive comparison between role-playing and persona-based dialogue refer to \cite{tseng-etal-2024-two}. Our work focuses on enabling role-playing LLMs to produce character-faithful responses. 

\textbf{Retrieval-Augmented Generation (RAG)} enhances LLMs by retrieving external knowledge to support informed, accurate, and contextually grounded responses \cite{lewis2020retrieval,liu-etal-2022-makes,zhuang2023toolqa,li-etal-2024-know}. Standard RAG struggles with capturing complex inter-entity relationships across multiple chunks \cite{guo2024lightrag} and often fails on general queries requiring comprehensive understanding of large knowledge bases \cite{Edge2024}. To address these limitations, recent work \cite{Edge2024,10.1145/3677052.3698671,wu2024medical,guo2024lightrag} leverages LLMs to construct knowledge graphs (KGs), where nodes represent entity attributes and edges encode inter-entity relationships. 

However, knowledge graph–enhanced methods overlook the entity ambiguity issue, where multiple names refer to the same character, and their retrieval process typically ignores the character's knowledge boundary, leading to responses that go beyond the intended role and produce out-of-character content. 
\section{RoleRAG}

\begin{figure*}[ht]
    \centering
    \includegraphics[width=1.05\linewidth]{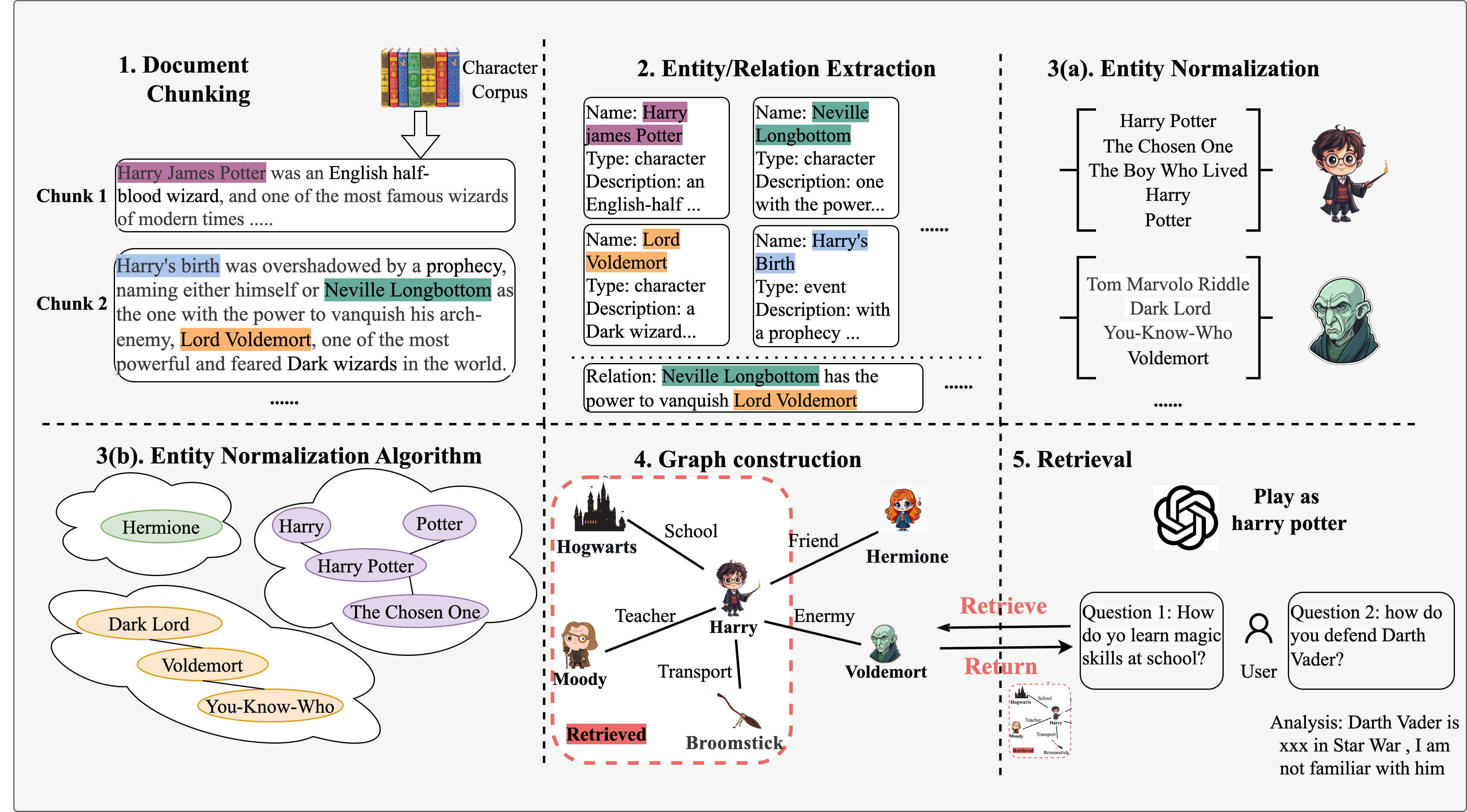}
    \caption{Workflow of our proposed RoleRAG.}
    \label{fig:workflow}
\end{figure*}

Our overall framework for RoleRAG is illustrated in Figure~\ref{fig:workflow} and consists of two novel modules specifically designed for the role-playing task: (1) an entity normalization module that removes semantically duplicated entities, and (2) a retrieval module that fetches question-relevant information while rejecting out-of-scope queries. 

\subsection{Entity and Relation Extraction} 
In role-playing, character corpora often originate from novels, TV series, or biographies, with descriptions that exceed LLM token limits.  Following prior work~\cite{Edge2024, wu2024medical, guo2024lightrag}, we split descriptions into chunks $\{\mathcal{D}_1, \mathcal{D}_2, ..., \mathcal{D}_n\}$, process them independently, and aggregate the results into a unified character profile.

For each chunk $\mathcal{D}_i$, we employ LLMs to meticulously extract entities, adhering to a predefined data structure: \{\textit{name}, \textit{type}, \textit{description}\}, denoted by $\mathbf{n}_i$. Furthermore, we prompt LLMs to identify structural relations between two entities, specifically, \{\textit{source}, \textit{target}, \textit{description}, \textit{strength}\}, denoted by $\mathbf{r}_i$, where \textit{description} and \textit{strength} denote the textual relationship and its intensity between the source and target nodes, respectively. After all chunks are processed, all entities and relations are stored in global databases $\mathcal{N}$ and $\mathcal{R}$.  

To enable semantic retrieval, each entity $\mathbf{n}_i$ is encoded into a high-dimensional vector $\mathbf{v}_i$ using a text embedding model applied to both its name and description. The resulting pairs ${\mathbf{n}_i, \mathbf{v}_i}$ are stored in a vector database $\mathcal{V}$ for efficient similarity-based retrieval. We define the retrieval interface as $f_k(\mathcal{V}, \mathbf{n})$, which returns the top $k$ entities most similar to a query entity.

\subsection{Entity Normalization}
\begin{algorithm}[t]
\caption{Entity Normalization Algorithm}
\label{alg:entity_normal}
\begin{algorithmic}[1]
\Require Entity Database $\mathcal{N}$.
\Ensure a unified name for each name group.
\State Initialize empty entity graph $\mathcal{G}$.
\State Initialize empty vector database $\mathcal{V}$.
\For{$\mathbf{n}_i \in \mathcal{N}$}
        \If{$\mathbf{n}_i \in \mathcal{V}$}
            \State {continue}; \Comment{node exists}
        \Else
            \State $\mathcal{N}_k = f_k(\mathbf{n}_i, \mathcal{V})$
            \State Insert $\mathbf{n}_i$ to $\mathcal{V}$
            \State Insert $\mathbf{n}_i$ to $\mathcal{G}$
        \EndIf 
        \For {$\mathbf{n}_j \in \mathcal{N}_k$}
            \If {$\mathbf{n}_i == \mathbf{n}_j$} \Comment{LLM prompt}
            \State Insert $\mathbf{n}_j$ to $\mathcal{G}$
            \State Connect $\mathbf{n}_i$ and  $\mathbf{n}_j$ in $\mathcal{G}$
            \Else
            \State continue
            \EndIf
        \EndFor 
\EndFor 
\State Count the number of connected components in $\mathcal{G}$
\For {each connected components $G$ in $\mathcal{G}$}
\State Select the unified name in $G$ \Comment{LLM prompt}
\EndFor
\end{algorithmic}
\end{algorithm}

To mitigate entity ambiguity, we introduce a semantic entity normalization procedure, detailed in Algorithm~\ref{alg:entity_normal}. Given all extracted entities, our algorithm iterates through each entity, retrieving the $k$ most semantically similar entities from the entity vector database. Next, we present these entity pairs, along with their names and descriptions, to the LLM, prompting it to determine if they refer to the same character. If the LLM identifies two entities as the same individual, we connect their corresponding nodes with an edge in the entity graph $\mathcal{G}$. After processing all entities, we partition the entity graph into multiple connected subgraphs, each representing a distinct individual, as illustrated in Figure~\ref{fig:workflow}. Finally, we prompt the LLM once more to generate a unified canonical name for each connected subgraph.

Compared to brute-force LLM-based pairwise comparisons, our method reduces LLM calls by a factor of $|\mathcal{N}|/k$, where $|\mathcal{N}|$ is the total number of entities and $k$ the number of entities retrieved from the vector database. We further leverage modern vector embedding techniques to accelerate retrieval by reducing the number of semantic dissimilar entities.

\subsection{Graph Construction} 
After identifying entity groups referring to the same character and assigning each group a unified canonical name, we construct a mapping table linking source names to their canonical forms. Subsequently, we normalize all raw names across both entity and relationship databases to facilitate effective retrieval. Since normalization reduces duplicate entities and relationships in $\mathcal{N}$ and $\mathcal{R}$, we summarize their descriptions using LLMs to preserve contextual details. 

Finally, we formally construct the knowledge graph from character database as follows,
\begin{align}
    \mathcal{\hat{G}} = \{\mathcal{\hat{N}}, \mathcal{\hat{R}}\}
\end{align}
where $\mathcal{\hat{N}}, \mathcal{\hat{R}}$ denote nodes and relationships after de-duplication. 

\subsection{Retrieval Module for Role-playing}
Given a user query, we first prompt an LLM to infer hypothetical contexts relevant to the desired response, inspired by HyDE~\cite{gao-etal-2023-precise}. Subsequently, we prompt the LLM with character profiles summarized from our knowledge graph to identify entities appearing in both the original query and the inferred hypothetical context. For each entity, the LLM returns its \textit{name}, \textit{entity type}, \textit{relevance to the designated character} (along with the underlying rationale), and \textit{specificity level} (either specific or general). Leveraging this information, we develop three distinct retrieval strategies to gather contextually appropriate content from the knowledge graph, supplementing the character summary provided to the LLM: 
\begin{itemize}[topsep=0pt, itemsep=0pt,leftmargin=*] 
\item For entities identified as outside the character's knowledge scope (e.g., querying an ancient figure about Apollo 11), we explicitly inform the LLM of their irrelevance along with the underlying rationale, thereby discouraging the LLM from providing hallucinatory responses. 
\item For specific entities, we first retrieve the top semantically similar entities from the vector database $\mathcal{V}$ based on the entity embeddings. Subsequently, we extract detailed descriptions of these entities and their relationships with the designated character from the knowledge graph to form the context.
\item For general entities (e.g., interests, hobbies), we retrieve entities from the 1-hop neighborhood of the target character, filtering out irrelevant entities based on their types. Descriptions of the remaining entities are then used to provide contextual details for response generation. 
\end{itemize}
Our retrieval strategy not only enriches character-related responses with detailed knowledge but also rejects out-of-scope questions that exceed the character’s cognitive boundaries, thereby enhancing knowledge exposure and reducing hallucinations in role-playing. 
\section{Experimental Setup}

\subsection{Baselines}

We compare RoleRAG against the following set of baselines: \textbf{Vanilla}, it prompts an LLM to role-play as a character with task description; \textbf{RAG}~\cite{lewis2020retrieval} retrieves chunks most semantically similar to a user query and provides them as context for LLM-based response generation; \textbf{Character profile} \cite{zhou-etal-2024-characterglm}, which provides the LLM with a profile of the character that the LLM is portraying; \textbf{GraphRAG}~\cite{Edge2024} retrieves relevant information from an indexed entity-relation knowledge graph.

We collect source materials from Wikipedia, Baidu Baike, and novels to construct the retrieval databases for both RAG and RoleRAG. For character profiles, we prompt GPT-4 to summarize the corresponding Wikipedia or Baidu Baike biography into a short paragraph, which is prepended to user queries to provide background context. 

\begin{figure*}
    \centering
    \includegraphics[width=0.8\linewidth]{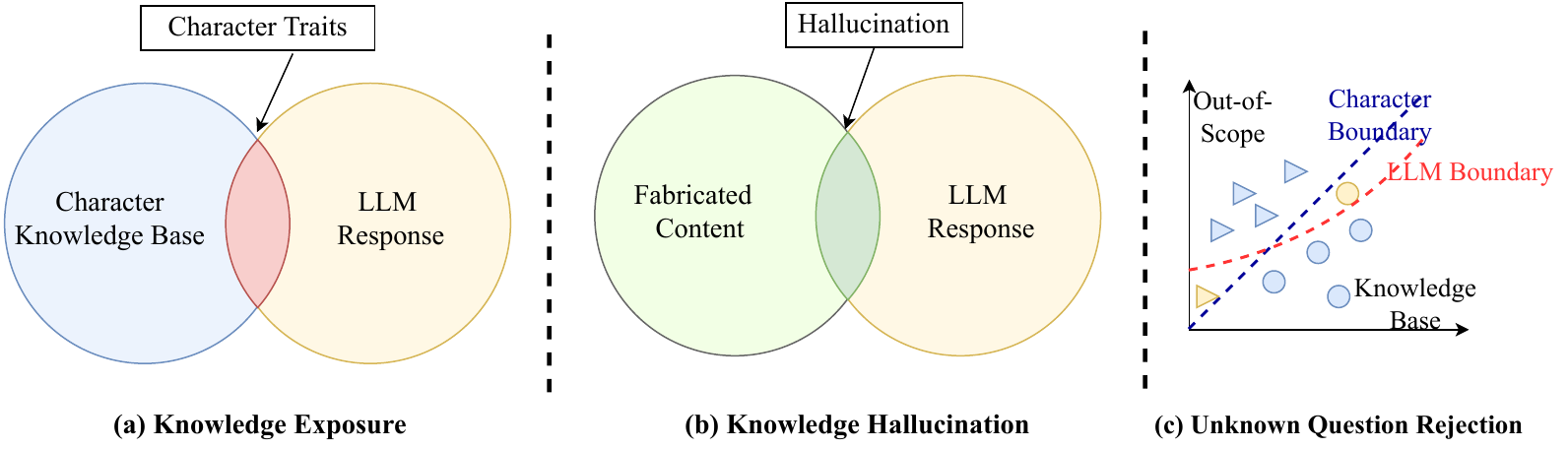}
    \caption{Illustration of evaluation metrics. We encourage LLMs to exhibit more personal traits, minimize fabricated content, and align more closely with the boundaries of character cognition.}
    \label{fig:evaluation}
    \vspace{-0.3cm}
\end{figure*}

\subsection{Evaluation Metrics}
Role-play LLMs should consistently embody the target role, provide accurate responses, maintain character integrity, and avoid factual errors. Following existing studies \cite{tu-etal-2024-charactereval,lu-etal-2024-large}, we perform our evaluation with the following metrics in Figure \ref{fig:evaluation}.  

\textit{Knowledge Exposure} measures the extent to which personalized traits—such as background, behavior, knowledge, and experiences—are accurately recalled from the character profile. \textit{Knowledge Hallucination} evaluates the precision of responses, focusing on the model's ability to avoid generating incorrect, misleading, or out-of-scope information. This is essential for maintaining the credibility and consistency of the LLM within the designed role. \textit{Unknown Questions Rejection} measures the model's self-awareness in role-playing by assessing its ability to recognize and communicate the boundaries of the character's knowledge.

To quantitatively evaluate these metrics, we follow prior work~\cite{, shao-etal-2023-character, dai2024mmrole, lu-etal-2024-large, wang-etal-2024-rolellm} and employ GPT-4o as a judge \cite{zheng2023judging} to rate the responses. We prompt GPT-4o to rate knowledge exposure and hallucination on a 1–10 scale. A higher knowledge exposure score indicates that the LLM demonstrates deep understanding of the character, while a lower hallucination score reflects responses free from misinformation about the character’s background. For self-awareness measurement, we prompt the LLM to assign a score of 1 if the response adheres to the character’s cognitive scope, and 0 otherwise. Since judge LLMs may exhibit biases during evaluation—such as the ``self-enhancement bias''~\cite{zheng2023judging}—we include human evaluators in the loop to verify and correct the scores produced by the judge LLM.  The detailed evaluation process is described in Appendix section \ref{apd:evaluation}. 

\subsection{Datasets}
To evaluate performance of our RoleRAG framework, we conducted experiments on three role-playing datasets: (1) \textbf{Harry Potter Dataset}, collected by us, this dataset contains seven characters from the Harry Potter series. Each character is presented with 20 role-specific questions (10 general questions about their interests and values, as well as 10 detailed questions about their experiences and relationships with others). (2)  \textbf{RoleBench-zh}, a subset of the RoleBench evaluation, this dataset includes five historical and fictional Chinese characters. This dataset contains both role-related and out-of-scope questions, 357 in total. For example, it includes a question about Apollo 11 directed at an ancient figure. (3) \textbf{Character-LLM} \cite{shao-etal-2023-character}, contains 859 questions, including role-related and out-of-scope questions. The statistics of the three datasets are provided in Appendix \ref{sec:appendix_dataset_stats}.

Our experiments are conducted on relatively small datasets featuring well-known characters or those from famous novels to ensure that details can be easily verified by human evaluators.

\subsection{Implementation Details}
In RoleRAG, we split the character profile into chunks of 600 tokens with an overlap of 100 tokens. GPT-4o mini is used as the LLM to extract entities and their relationships, perform entity normalization, and merge descriptions of duplicate entities. We use OpenAI's ``text-embedding-3-large model'' to encode entity descriptions into vector representations with an embedding dimension of 3,072. Cosine distance is used to measure the similarity between entities. 

To assess RoleRAG's usability, we perform experiments with various LLMs, including open-source LLMs (including Mistral-Small 22b \cite{mistralsmall}, Llama3.1 8b, Llama3.3 70b \cite{dubey2024llama}, Qwen2.5 14b \cite{yang2024qwen2}), proprietary LLMs (OpenAI GPT series \cite{gpt4o}), and LLMs specifically tailored for role-playing tasks (Doubao Pro 32k\footnote{https://www.volcengine.com/product/doubao}). 
\section{Experimental Results}

\subsection{Main Results}

\begin{table*}[th]
\caption{Our main experimental results on the Harry Potter, RoleBench-zh, and CharacterLLM datasets. The reported scores are the average across all questions in each dataset, and $\uparrow$ / $\downarrow$ means higher/lower results are better. Human evaluators are recruited to verify and correct GPT-4o's score.}
\small
\centering
\label{tab:main_results}
\begin{tabular}{l|l|lll|lll|lll}
\toprule
\multirow{2}{*}{Model}               & \multirow{2}{*}{Method} & \multicolumn{3}{c|}{Harry Potter}                                            & \multicolumn{3}{c|}{RoleBench-zh}                                            & \multicolumn{3}{c}{CharacterLLM \ddag}              

\\                                   &                         & \multicolumn{1}{c}{KE $\uparrow$} & \multicolumn{1}{c}{KH $\downarrow$} & \multicolumn{1}{c|}{UQR $\uparrow$} & \multicolumn{1}{c}{KE $\uparrow$} & \multicolumn{1}{c}{KH $\downarrow$} & \multicolumn{1}{c|}{UQR $\uparrow$} & \multicolumn{1}{c}{KE $\uparrow$} & \multicolumn{1}{c}{KH $\downarrow$} & \multicolumn{1}{c}{UQR $\uparrow$} \\ \midrule
\multicolumn{11}{c}{Open-source General Models} \\ \midrule
\multirow{5}{*}{Mistral-Small (22b)} & Vanilla  & 7.457 & 2.229  & --- & 4.398  &  5.731 & 0.510  & 8.535 & 1.794 & 0.894   \\  
 & RAG  & \textbf{7.786} & 2.486  & --- & 4.905  & 5.367 & 0.580  & 8.871 & 1.538 &  0.929  \\  
 & User profile  & 7.650 &  2.293 & --- & 5.182  & 3.890 & \textbf{0.711}  & 8.861 &1.570  &  0.932  \\  
  & GraphRAG  &7.356& 2.488 & --- & 5.328 & 4.459 &  0.613 & 8.963 &1.572  &  0.925  \\  
 & RoleRAG  & 7.550 &  \textbf{2.150} & --- &  \textbf{5.585} & \textbf{3.961} &  0.678 & \textbf{9.057} & \textbf{1.404} & \textbf{0.959}   \\       \midrule                 
\multirow{5}{*}{Llama 3.1 (8b)} & Vanilla  & 7.579 & \textbf{2.200} & --- &  4.115 & 6.232 & 0.462  & 7.932 & 2.613 & 0.819   \\  
 & RAG  & 7.486 & 3.214 & --- & 4.728  & 5.389 & 0.600  & 8.505 & 2.084 &  0.884  \\  
 & User profile  & 7.057 & 3.657  & --- & 5.047  & 4.843 & 0.569  & 8.292 & 2.174 &  0.875  \\  
& GraphRAG  & 7.373 & 2.833 & --- & 5.479  &  4.367 &  \textbf{0.678} & 8.543 & 2.019 &  0.900  \\
 & RoleRAG  & \textbf{7.750} & 2.352 & --- & \textbf{5.608}  & \textbf{4.126} & 0.661  & \textbf{8.653} & \textbf{1.961} &   \textbf{0.908} \\       \midrule                                    
\multirow{5}{*}{Qwen 2.5 (14b)} & Vanilla  & 7.614 & 2.129 & --- &  6.238 & 3.352 &  0.734 & 8.709 & 1.656 &  0.907  \\  
 & RAG  & 7.707 & 2.371 & --- & 6.583   &  3.020 & 0.773 & 9.067 & 1.356 & 0.959   \\  
 & User profile  & 7.764 & 2.693 & --- & 6.605  & 3.020 &  0.818 & 9.039 & 1.382 &  0.953  \\ 
 & GraphRAG  & 7.762  & 2.433  & --- & 6.686   & 2.888 & 0.790   & 9.230 & 1.321 &  0.956  \\ 
 & RoleRAG  & \textbf{7.986} & \textbf{2.071}  & --- & \textbf{6.798}  & \textbf{2.538} &  \textbf{0.832} & \textbf{9.238} & \textbf{1.231} &  \textbf{0.974}  \\       \midrule               
\multirow{5}{*}{Llama3.3 (70b)} & Vanilla  & 7.414 & 2.279  & --- &  6.034 & 3.709 & 0.689  & 8.811 & 1.419 &  0.929  \\  
 & RAG  & 8.243 & 2.071  & --- & 6.031  & 3.546 & 0.751  &  9.198& 1.352 &  0.962  \\  
 & User profile  & 8.021 & 2.050 & --- & 6.457  & 3.014 & 0.754  &9.258  & 1.272 &  0.964  \\ 
 & GraphRAG  & 8.352  & 2.070 & --- & 6.092  & 3.521  & 0.714  &\textbf{9.302}  & 1.275 &  0.967  \\ 
 & RoleRAG  & \textbf{8.564} & \textbf{1.743} & --- & \textbf{6.723}  & \textbf{2.622} &  \textbf{0.837} & 9.270 & \textbf{1.265} &  \textbf{0.974}  \\       \midrule  
\multicolumn{11}{c}{Close-source General Model} \\ \midrule
\multirow{5}{*}{GPT-4o-mini } & Vanilla  & 7.643 & 2.121  & --- & 5.863  &4.202  & 0.714  & 8.789  & 1.492 &  0.925  \\  
 & RAG  & 8.493 &  1.750 & --- & 5.986  & 3.930 &  0.709 & 8.996 & 1.311 &  0.954  \\  
 & User profile  & 8.221 & 2.021  & --- & 6.232  &3.754  & 0.733  & 9.009 & 1.317 &  0.945  \\  
 & GraphRAG  &8.729 & 1.776  & --- & 6.445  & 3.429 & 0.717  & 9.136 & 1.308 &  0.958  \\  
 & RoleRAG  & \textbf{8.821} & \textbf{1.571}  & --- & \textbf{6.994}  & \textbf{2.697}  &  \textbf{0.857} & \textbf{9.138} & \textbf{1.211} &  \textbf{0.978} \\       \midrule                                       
\multicolumn{11}{c}{Close-source Role-playing Model} \\ \midrule
\multirow{4}{*}{Doubao Pro 32K} & Vanilla  & 7.193 &  2.257 & --- &6.840   &  3.745&  0.860 & 8.522 & 1.639 & 0.891   \\  
& RAG  & 8.179 & 1.814  & --- &  7.170 & {2.246} & 0.880  & 8.836 & 1.379 &  0.939  \\  
& User profile  & 7.450 & 2.179  & --- & 7.207   & 2.429 & {0.905}  & 8.927 & 1.351 &  0.932  \\  
& GraphRAG & 8.040& 1.780  & --- &  6.866  & 2.087 & 0.902  & 8.929 & 1.361 &  0.932  \\  
& RoleRAG  & \textbf{8.221} & \textbf{1.564} & --- & \textbf{7.733}  & \textbf{1.689 }& \textbf{0.952}  & \textbf{8.970} & \textbf{1.313} &  \textbf{0.956}  \\       \bottomrule  
                                     
\multicolumn{11}{r} \# {KE:} Know exposure [0, 10], {KH:} Knowledge hallucination [0, 10], {UQR:} Unknown question rejection \{0, 1\}.             \\
\multicolumn{11}{r} \ddag Human evaluation takes extremely longer on this dataset, we average scores from two trials of GPT4o.   
\end{tabular}%
\vspace{-0.3cm}
\end{table*}

Our main results are shown in Table \ref{tab:main_results}. Overall, the results show that RoleRAG performs better than the baseline methods. In many instances, a smaller LLM with RoleRAG, e.g., Qwen 2.5 (14b), can outperform larger LLMs, e.g., Llama 3.3 (70b), without it, demonstrating the effectiveness of RoleRAG. While adding character background improves knowledge exposure and reduces hallucination compared to vanilla approaches, RoleRAG outperforms other retrieval-based baselines by structuring information for efficient access to character details and relationships, enabling more accurate role-playing. For unknown questions, RoleRAG outperforms baseline methods, even when those are explicitly instructed not to answer out-of-scope queries. We attribute this to RoleRAG’s relevance analysis during retrieval, along with rationale generation, which helps prevent implausible responses—such as asking Harry Potter about events in Star Wars.

Fine-tuning LLMs for role-playing can improve performance, as shown by Doubao Pro on the RoleBench-zh dataset. However, the vast number of characters makes it impractical to fine-tune models for all possible roles. Additionally, defining and enforcing cognitive boundaries during fine-tuning remains a challenging, unsolved problem. These limitations are evident in Doubao Pro’s weaker performance on the Harry Potter and CharacterLLM datasets, along with its lower self-awareness. In contrast, RoleRAG enables both general-purpose and fine-tuned LLMs to access character-specific knowledge effectively. 

The results in Table~\ref{tab:main_results} appear only marginally improved due to the judge LLM’s tendency to assign high knowledge exposure scores and low hallucination scores when responses lack major errors. For example, scores of 8–9 are often given for generally appropriate answers, while human evaluators tend to adjust scores only in cases of significant faults rather than making fine-grained changes. As a result, the high baseline scores from LLM judges leave limited room for observable improvement.

\begin{table}[t]
\centering
\caption{Ablation studies on RoleBench-zh datasets.}
\label{tab:ablation}
\resizebox{0.9\linewidth}{!}{
\begin{tabular}{l|l| c|c |c} 
\toprule
Entity Normalization  & Retrieval & KE & KH & UQR  \\ \midrule
Without  & Local  search & 6.006 &4.126	&0.745\\ \midrule
With & Local  search & 6.431 &3.409 &0.770\\ \midrule
Without & Our retrieval & 6.154	& 3.454	& 0.762 \\ \midrule
With & Our retrieval &6.994&2.697&	0.857\\ \bottomrule
\end{tabular}}

\caption{Performance of RoleRAG on general questions on Harry Potter dataset.}
\label{tab:general}
\resizebox{\linewidth}{!}{%
\begin{tabular}{l| c c|cc}
\toprule
\multirow{2}{*}{Model}  &  \multicolumn{2}{c}{KE}  & \multicolumn{2}{c}{KH}\\
& Vanilla & RoleRAG & Vanilla  & RoleRAG\\ \midrule
Mistral-Small (22b) & 7.486 &7.685& 1.457 & 1.485\\
Llama3.1 (8b) & 7.714 & 8.342 &1.343 & 1.614\\
Qwen 2.5 (14b) & 7.614 & 8.157 & 1.414 & 1.371 \\
Llama 3.3 (70b) & 7.414 & 8.814 & 1.557 & 1.086 \\
GPT-4o mini & 7.671 & 8.957 & 1.371 & 1.157 \\
Doubao Pro 32K & 7.300 & 8.414 & 1.586 &  1.057\\ \bottomrule
\end{tabular}}

\caption{Performance of RoleRAG on specific questions on Harry Potter dataset.}
\label{tab:specific}
\resizebox{\linewidth}{!}{%
\begin{tabular}{l| c c|cc}
\toprule
\multirow{2}{*}{Model}  &  \multicolumn{2}{c}{KE}  & \multicolumn{2}{c}{KH}\\
& Vanilla & RoleRAG & Vanilla  & RoleRAG\\ \midrule
Mistral-Small (22b) & 6.587 & 7.414 & 2.6 & 2.814\\
Llama3.1 (8b) & 6.842 & 7.157 & 3.058 & 3.070\\
Qwen 2.5 (14b) & 7.425 & 7.902 & 2.842 & 2.771\\
Llama 3.3 (70b) & 7.213 & 8.314& 3.000 &2.400\\
GPT-4o mini & 7.314 & 8.686 & 2.871 & 1.986 \\
Doubao Pro 32K &7.085 & 8.029 & 2.929 & 2.071 \\ \bottomrule
\end{tabular}}

\caption{Performance of RoleRAG across characters with varying frequencies in the Harry Potter series, listed from highest to lowest frequency. }
\label{tab:frequency}
\resizebox{\linewidth}{!}{%
\begin{tabular}{l| c c|cc}
\toprule
\multirow{2}{*}{Model}  &  \multicolumn{2}{c}{KE}  & \multicolumn{2}{c}{KH}\\
& Vanilla & RoleRAG & Vanilla  & RoleRAG\\ \midrule
Harry Potter & 7.77 & $8.11_{+0.34}$ & 1.69 & $1.97_{+0.28}$ \\
Hermione Granger & 7.57 & $8.23_{+0.66}$ & 2.58 & $2.28_{-0.3}$\\
Voldemort & 7.99 & $8.37_{+0.38}$ & 1.85 & $1.98_{+0.13}$\\
Alastor Moody & 7.47& $7.83_{+0.36}$ & 2.77 & $2.63_{-0.14}$ \\
Ludovic Bagman & 7.08 & $8.18_{+1.1}$ & 2.46  & $1.68_{-0.78}$\\
Padma Patil & 7.14& $8.4_{+1.26}$ & 2.21 & $1.34_{-0.87}$\\ 
Roger Davies & 7.24 & $7.94_{+0.7}$ & 2.08 & $1.83_{-0.25}$ \\\bottomrule
\end{tabular}}
\end{table}

\subsection{Ablation Studies}

Note that our method is built upon knowledge graph enhanced retrieval. Different from GraphRAG, we introduce the entity normalization to merge duplicated entities during graph construction and a retrieval strategy for role-playing. In this ablation study, we disable entity normalization and adopt the local search that starts from the most similar nodes from query embedding, and expanding through its neighborhood and community in GraphRAG. The experiment results are illustrated in Table ~\ref{tab:ablation}, we can see that: 1) the most significant improvement comes from the combination of RoleRAG and the novel retrieval strategy; 2) the retrieval method could clearly enhance the boundary awareness by providing relevance to the character. 

\subsection{RoleRAG for General Questions}

Table~\ref{tab:general} presents knowledge exposure and hallucination scores for general questions in the Harry Potter dataset. While LLMs show low hallucination, they reveal few character-specific traits. We hypothesize that LLMs have internalized general knowledge from large-scale pretraining but lack role-specific details. In our RoleRAG, we retrieve 1-hop neighbors of the character matching the type of general keywords, enriching the response with relevant context and significantly improving knowledge exposure while keeping low hallucination.

\subsection{RoleRAG for Specific Questions}
Table \ref{tab:specific} demonstrates knowledge exposure and hallucination scores for specific questions from the Harry Potter dataset. Compared with responses to general questions, when asked about details, LLMs tend to fabricate stories or are reluctant to provide specific information. With our RoleRAG, we observe a clear improvement in knowledge exposure and hallucination scores after retrieving detailed entity information mentioned in user questions from the knowledge base. We also observe an interesting phenomenon: smaller LLMs tend not to incorporate the retrieved knowledge into their responses as effectively as larger LLMs. 

\subsection{RoleRAG for Minority Groups}

Table~\ref{tab:frequency} reports performance across characters in the Harry Potter series, sorted by their frequency of appearance. The results demonstrate that for popular characters like `Harry Potter', LLMs exhibit higher knowledge exposure and lower hallucination rates. Conversely, less commonly mentioned characters tend to show reduced knowledge accuracy and increased instances of fabricated content. These results show that with the aid of RoleRAG, characters that appear less frequently, such as `Ludovic Bagman' and `Padma Patil', benefit significantly in terms of enhanced knowledge exposure and reduced fabrication of content.

\subsection{RoleRAG for Out-of-scope Questions} 

\begin{figure}[t]
    \centering
    \includegraphics[width=0.9\linewidth]{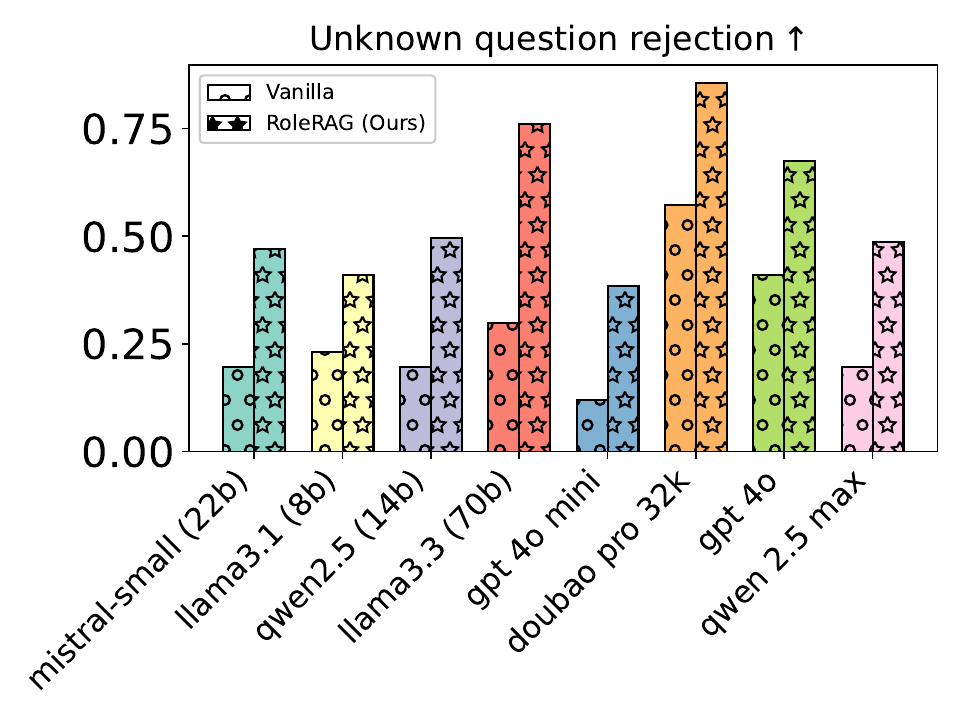}
    \caption{Experiments of out-of-scope questions in RoleBench-zh dataset.}
    \label{fig:out_of_scope_experiments}
    \vspace{-0.5cm}
\end{figure}

Figure~\ref{fig:out_of_scope_experiments} shows that when role-playing, LLMs tend to answer all questions—even those beyond the character’s knowledge scope. This suggests that LLMs often fail to fully adopt the perspective of the target character, instead relying on their internalized knowledge—an issue observed even in larger models like GPT-4o and Qwen2.5-Max. While the strong performance of Doubao Pro shows that fine-tuning can improve awareness of a character’s cognitive boundary, it lacks adaptability to new characters without task-specific data. Overall, regardless of model size or fine-tuning, the results demonstrate that RoleRAG equips LLMs with the information needed to correctly reject out-of-scope questions, better aligning their cognitive boundaries with the intended character.
\section{Conclusion}
When tasked with role-playing, LLMs often generate responses that lack depth in character knowledge and introduce information outside the character’s known universe—a role-specific form of hallucination. To address these issues, in this paper, we introduce RoleRAG, a novel framework for role-playing that merges duplicated entities and enhances the retrieval of relevant information. Additionally, our retrieval module assesses entity relevance to the target character, enabling accurate content generation while effectively rejecting unrelated questions. Through rigorous experimentation, we demonstrated that RoleRAG consistently outperforms relevant baselines. The success of RoleRAG highlights its potential as a powerful tool for improving the reliability and authenticity of role-playing models, paving the way for more sophisticated, context-aware conversational agents in a variety of applications. 
\section{Limitations}

A minor concern in our work is the evaluation of the responses generated by LLMs. It is difficult to recruit human evaluators who have deep knowledge about the characters and stories used in our evaluations. Even if evaluators are familiar with the characters and stories, they may need more detailed information to accurately judge whether a generated response is sensible and does not contain hallucination. Therefore, we use LLMs as evaluators in our experiments, then verified by human annotators. However, we observed that LLMs tend to assign over-confident scores, which can mislead human evaluators and render the scores insufficiently discriminative in our experiments.

A possible direction to explore is how to prompt an LLM to recognize and understand the limits of character knowledge when engaged in role-play. Given that LLMs are trained on massive, diverse datasets, they often possess knowledge far beyond what the characters they are asked to portray would realistically know. As a result, managing these knowledge boundaries becomes crucial to ensuring more authentic role-playing. Defining the scope and limits of a character's knowledge is not only necessary to prevent the model from introducing irrelevant or inaccurate information, but it also directly improves the accuracy of knowledge exposure within the context of the character. Ultimately, addressing this challenge could significantly enhance the believability and effectiveness of LLMs in role-playing scenarios, fostering more realistic and emerging interactions.

Another limitation of our work is that we focused on single-turn conversations. Multi-turn conversations present unique challenges, including maintaining consistency across turns, ensuring that the LLM remains in-character, and effectively managing the dialogue history. As multi-turn conversations often require the model to recall and build upon previous interactions, there is an increased risk of the model deviating from the character's personality or losing track of essential details. In the future, we plan to investigate how to address these challenges.

In retrieval-based methods, the quality of the response generated by an LLM depends on the model's ability to utilize the information retrieved. However, it is not fully understood how LLMs incorporate this retrieved knowledge into their responses. We have observed numerous instances where LLMs contradict the retrieved information. Thus, gaining a deeper understanding of the internal mechanisms of in-context learning is crucial to improving retrieval-based approaches.

\section{Ethics}
We will release our code base publicly as part of our commitment to the open source initiative. However, it is important to recognize that role-playing with these tools can lead to jailbreaking, and misuse may result in the generation of biased or harmful content, including incitement to hatred or the creation of divisive scenarios. We truly hope that this work will be used strictly for research purposes. 

With our proposed RoleRAG, we aim to effectively integrate role-specific knowledge and memory into LLMs. However, we must acknowledge that we cannot fully control how LLMs utilize this knowledge in dialogue generation, which could still result in harmful or malicious responses. In the future, we plan to investigate the mechanisms of prompting to more deliberately control response generation. Additionally, it is crucial to scrutinize responses in high-stakes and sensitive scenarios to ensure safety and appropriateness.

\bibliography{custom}

\appendix
\clearpage
\section{Comparison of LLM-based Role-playing approaches}
\label{sec:appendix_a}
Table \ref{tab:method_summary} shows a comparison of different methods used for using LLMs in role-playing tasks. Fine-tuning–based approaches require extensive data collection and are computationally expensive, and they often fail to generalize to roles beyond the training corpus, as each character has a distinct knowledge. Moreover, LLMs inherently encode vast general knowledge, which they may draw upon when answering queries—often leading to fabricated or out-of-scope content. Defining clear character boundaries remains a challenge for fine-tuning–based approaches. Retrieval-based methods eliminate the need for model training and costly data labeling. However, their effectiveness depends on efficiently retrieving query-relevant context from a large character knowledge base through a robust indexing system.  

\begin{table*}[htbp]
    \centering
    \caption{Comparison of different LLM role-playing approaches.}
   \scalebox{0.8}{
   \begin{tabular}{l c cc}
        \toprule
       Methods  & Fine-tuning Based  & Retrieval Based  &  Plugin Model\\ \midrule
       LLM Training  & YES & No & YES \\
       Character Data Labeling  & YES & No & YES \\
       Computational Burden & High & Low &  Moderate  \\ 
       Character Data Organization & \makecell{All characters\\ shared} & \makecell{One character,\\ one corpus} &  \makecell{One character,\\ one plugin} \\ 
       Adaptation to Unseen Roles & Hard & Easy &  Hard  \\
       Modifying LLMs' Knowledge & Hard & Easy &  Hard  \\ \bottomrule
    \end{tabular}}
    \label{tab:method_summary}
\end{table*}

\section{Dataset Statistics}
\label{sec:appendix_dataset_stats}
The statistics of our experimental datasets are illustrated in Table \ref{tab:statictics}. 
In our experiment, recruiting evaluators who can recall the complete knowledge base of a specific character is challenging, and web searches are often required during evaluation. For instance, assessing a batch of 357 response in the RoleBench-Zh dataset takes approximately \textbf{three hours} per evaluation session; The cost of evaluating LLM generation of CharacterLLM dataset with GPT-4 is approximately $5$ US dollars.

\begin{table}[H]
    \centering
    \caption{Statistics of the experimental datasets.}
    \begin{tabular}{l c cc}
        \toprule
       Datasets  &\#Roles & In  & Out of\\
        & & Scope  & Scope\\ \midrule
       Harry Potter  & 7 & 140 & - \\
       RoleBench-Zh & 5 & 240 & 117 \\
       Character-LLM & 9 & 814 &  45  \\ \bottomrule
    \end{tabular}
    \label{tab:statictics}
\end{table}

\section{Evaluation Process}
\label{apd:evaluation}

To judge the generated responses according to the above metrics, we make use of GPT-4o to act as a judge LLM by rating the responses. Powerful LLMs such as GPT-4 have been widely employed as evaluators in recent studies \cite{shao-etal-2023-character,dai2024mmrole,lu-etal-2024-large,wang-etal-2024-rolellm} where GPT-4 is prompted to give scores for generated output on a defined scale, or to compare responses and select which one is better. However, there are some concerns about the reliability of LLMs to rate generated responses. Therefore, based on recent works that explore the use of LLMs as judges, we adopt a few measures to increase the reliability of the scores in our experiments. First, we prompt the LLM to generate an analysis before it scores the response. This approach follows recent research \cite{shen-etal-2023-large, zheng2023judging} and is based on the success of Chain-of-Thought prompting \cite{wei2022chain}. Following Ditto \cite{lu-etal-2024-large}, we set the temperature of GPT-4o to 0.2 to penalize creativity during evaluation.

To avoid biases that judge LLMs may have, such as the ``self-enhancement bias'' \cite{zheng2023judging}, we include humans in the evaluation process to verify the scores produced by the judge LLM. The human evaluator can use the analysis produced by the judge LLM, as well as any other information sources they want to use, to determine whether the score is sensible. The human evaluator can adjust the score if they feel that it is not correct. We use three different prompts to generate scores for each metric, which can be found in Appendix \ref{sec:appendix_prompts}. 

\begin{figure}[htbp]
  \centering
  \begin{minipage}[b]{0.48\textwidth}
    \includegraphics[width=\textwidth]{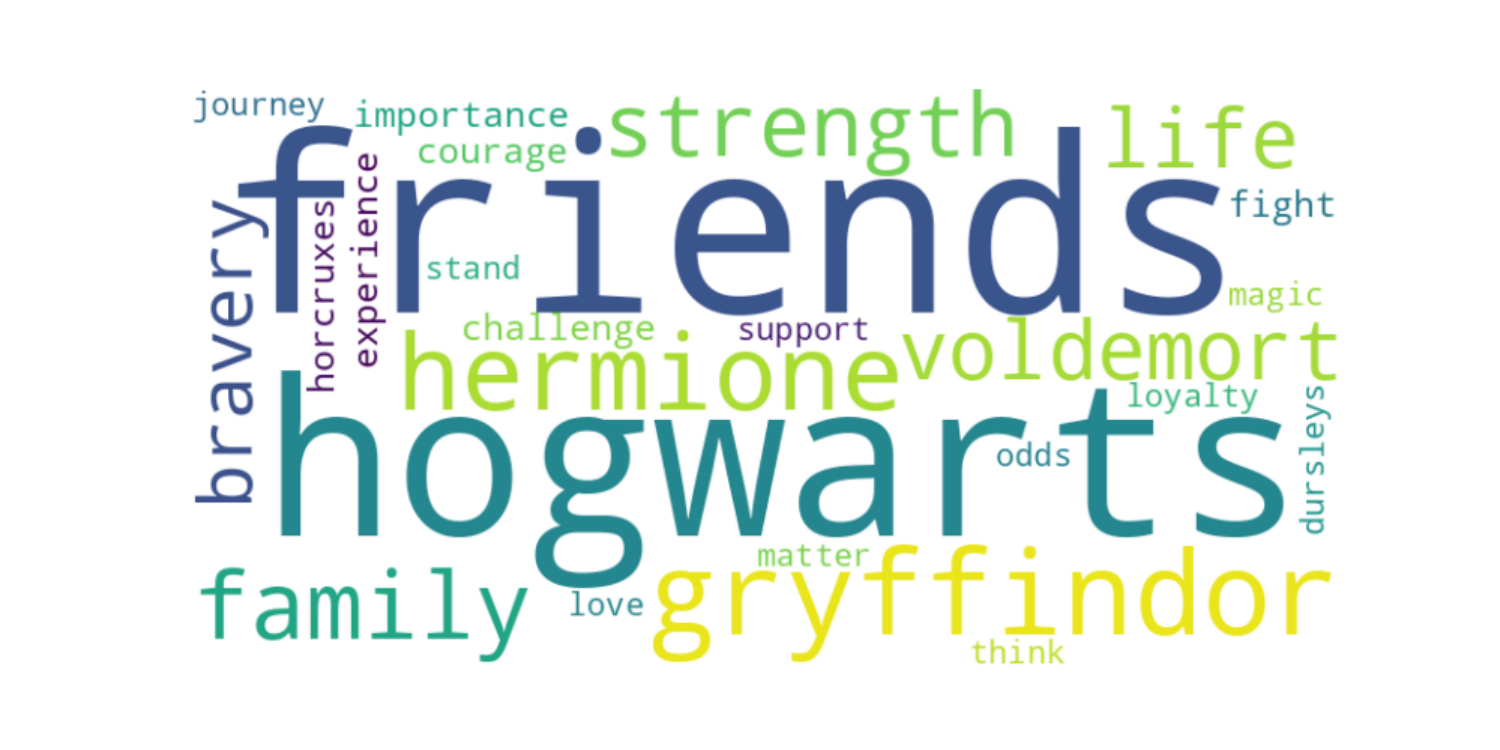}
    \caption{Word cloud for responses generated by GPT-4o mini when role-playing as Harry Potter.}
    \label{fig:word_cloud_harry}
  \end{minipage}
  \hfill
  \begin{minipage}[b]{0.48\textwidth}
    \includegraphics[width=\textwidth]{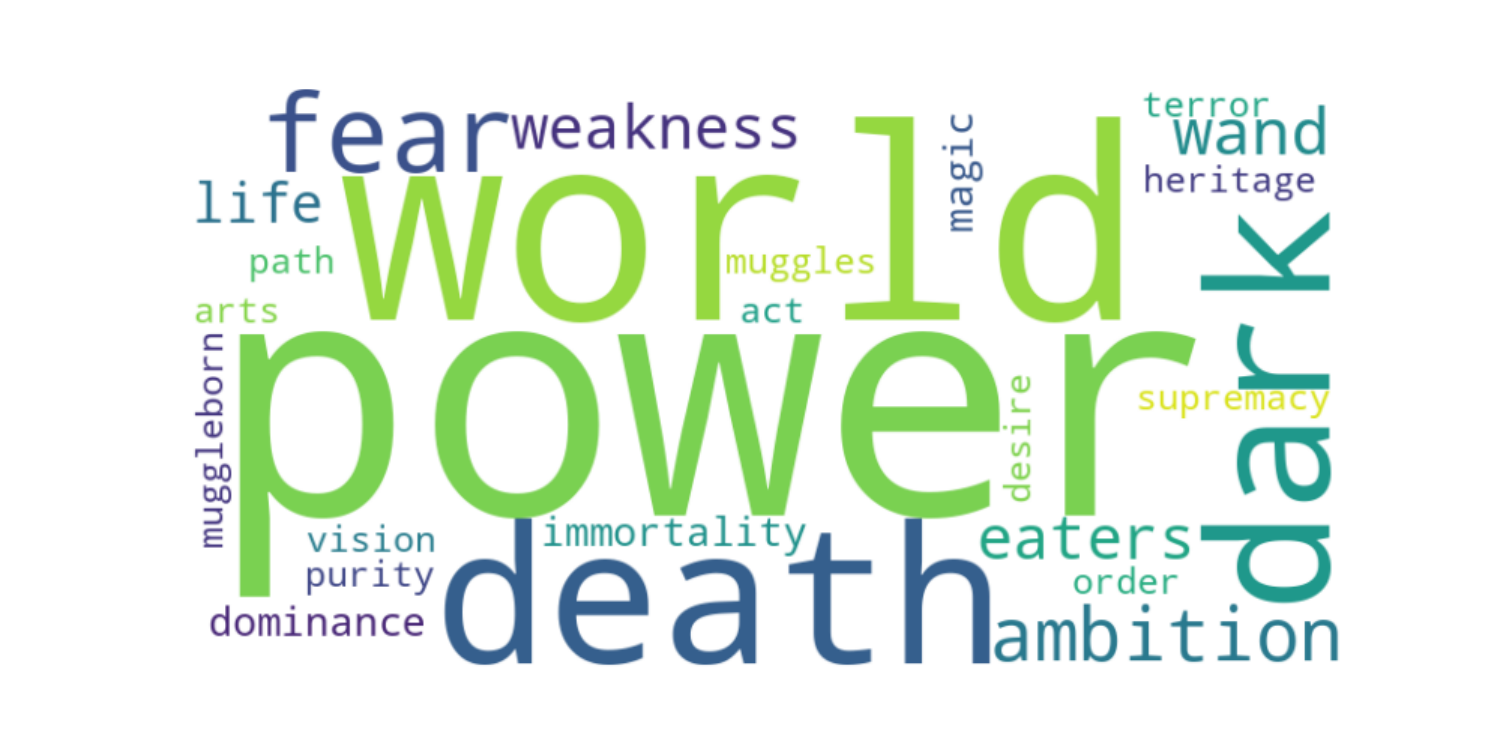}
    \caption{Word cloud for responses generated by GPT-4o mini when role-playing as Voldemort.}
    \label{fig:word_cloud_voldemort}
  \end{minipage}
\end{figure}

\section{Additional Experiments}
\subsection{Word Clouds}
\label{sec:word_clouds}
In this experiment, we collect all responses generated by GPT-4o-mini when role-playing characters from the Harry Potter series and visualize them using word clouds to highlight frequently used terms. Larger words in the figures indicate higher frequency. As shown in Figure~\ref{fig:word_cloud_harry}, when playing as Harry Potter, the model emphasizes terms such as friends, Hogwarts, Hermione, Voldemort, and brave. In contrast, Figure~\ref{fig:word_cloud_voldemort} shows that when role-playing Voldemort, the most frequent terms include world, power, death, and dark. 

\subsection{Demonstration of RoleRAG retrieval}

Figure~\ref{fig:demo} illustrates the types of information retrieved by RoleRAG in response to an interview question posed to an LLM role-playing Ludwig van Beethoven. The question asks about Beethoven’s relationship with Haydn and Mozart. RoleRAG first identifies the entities in the query, assesses their relevance to Beethoven, and determines their specificity. 

Since all three are specific entities related to Beethoven, we locate them in the knowledge graph using semantic matching and the name mapping table. We then retrieve detailed information about each entity and their relationships with Beethoven. Finally, the retrieved context is aggregated and provided to the LLM to answer the user’s question.

\begin{figure*}
    \centering
    \begin{tcolorbox}[
      colback=white, colframe=black, arc=3mm, width=1\linewidth,
      title=\textbf{Demonstration of RoleRAG retrieval step}, 
      coltitle=white, colbacktitle=gray, fonttitle=\bfseries
    ]
    \textbf{Question:}\{Beethoven, What was the nature of your relationship with Haydn and Mozart?\}
\vspace{5pt}
    
 \textbf{Entity Extraction:} \{entity type, name, familiarity with Beethoven (reason, YES|No), Level (specific|general)\}

(character<|>Beethoven<|>Beethoven must know himself<|>Yes<|>specific)

(character<|>Joseph Haydn<|>Haydn was Beethoven's teacher and a significant influence on his early compositions<|>Yes<|>specific) 

(character<|>Wolfgang Amadeus Mozart<|>Mozart was a contemporary of Beethoven and a major influence on his work, although they did not have a direct relationship<|>Yes<|>specific)

\vspace{8pt}
 
\textbf{Retrieved Content:} 

---- Entity Information-----

BEETHOVEN,CHARACTER, ``Ludwig van Beethoven was a renowned German composer and pianist who significantly contributed to the landscape of classical music, transitioning between the late Classical and early Romantic periods. Born in Bonn, he moved to Vienna, ... ''
\vspace{5pt}

JOSEPH HAYDN,CHARACTER,``Joseph Haydn was a prominent composer who played a significant role in Beethoven's musical education, particularly in the area of counterpoint. He was an esteemed figure in the Classical music genre and served as a pivotal influence during Beethoven's early years in Vienna, introducing him to various musical concepts and techniques.''

\vspace{5pt}
MOZART,CHARACTER,``Wolfgang Amadeus Mozart was a prominent and renowned composer whose significant success and legacy played a crucial role in influencing Johann, particularly in promoting Beethoven as a child prodigy. His impact extended deeply into Beethoven's formative years as a composer, shaping the young musician's development and artistic direction.''

\vspace{5pt}
---- Relation Information-----

BEETHOVEN,JOSEPH HAYDN, ``Beethoven, an influential composer in the history of classical music, studied under the guidance of Joseph Haydn, a prominent figure known for his significant contributions to the development of the symphony and string quartet. Under Haydn’s direction, Beethoven acquired essential skills in counterpoint and composition, which greatly influenced his early works. Additionally, Haydn played a crucial role in introducing Beethoven to a broader musical community and providing mentorship in Vienna, marking a pivotal development in Beethoven's career. This relationship between the two composers highlights the impact of mentorship and collaboration in the evolution of classical music.''

\vspace{5pt}
BEETHOVEN,MOZART,``Beethoven, a prominent composer, was notably influenced by the style of Mozart, which played a significant role in his artistic development. This relationship highlights the profound impact that Mozart's musical elements had on Beethoven's compositions.''
\end{tcolorbox}
\caption{Use case of retrieval step in our RoleRAG.}
    \label{fig:demo}
\end{figure*}



\section{Prompts in our experiments}
\definecolor{darkgreen}{RGB}{2, 191, 35}
\label{sec:appendix_prompts}
This section contains the prompts used in our experiments. Figures \ref{fig:exposure_prompt}, \ref{fig:hallucination_prompt}, \ref{fig:rejection_prompt},  and \ref{fig:generation_hp_prompt} show the prompts used for generation and scoring.  Green text in curly braces represent text that is replaced based on the context.

\begin{figure*}
    \centering
    \begin{tcolorbox}[
      colback=white, colframe=black, arc=3mm, width=1.1\linewidth,
      title=\textbf{Prompt for Generating Knowledge Exposure Scores}, 
      coltitle=white, colbacktitle=gray, fonttitle=\bfseries
    ]
    Play act as \textcolor{darkgreen}{\{character\}}, who is \textcolor{darkgreen}{\{description\}}.
    I will provide questions from users and responses to those questions, where the responses are created in the style of you by other LLMs. You are required to judge and assess whether the response to a user's question match the knowledge and experience of you. 
    To evaluate the response, consider the following aspects:

    \vspace{5pt}
    (1) Whether stories/events discussed occurs in the same period as you.
    
    (2) Whether objects in the response have relevance to you.
    
    (3) Whether locations in the response are correct in your experience.
    
    (4) Whether persons mentioned have accurate relationships with you.
    \vspace{8pt}
    
    Based on the given instructions, provide a brief analysis of the given response. Then rate the response using a single score from 1 to 10, where a higher score indicates greater consistency with your knowledge.
    \vspace{8pt}
    
    Please provide your output in the following format:
    
    Analysis: <analysis>
    
    Rating: <rating>
    
    \vspace{5pt}
    \#\#\#\#\#\# Test Begin \#\#\#\#\#\#
    
    \textbf{User Question:} \textcolor{darkgreen}{\{question\}}
    
    \textbf{Response:} \textcolor{darkgreen}{\{response\}}
    
    \textbf{Output:}
    
    \end{tcolorbox}
    \caption{The prompt used for generating knowledge exposure scores.}
    \label{fig:exposure_prompt}

     \begin{tcolorbox}[
      colback=white, colframe=black, arc=3mm, width=1.1\linewidth,
      title=\textbf{Prompt for Generating Knowledge Hallucination Scores}, 
      coltitle=white, colbacktitle=gray, fonttitle=\bfseries
    ]
    Play act as \textcolor{darkgreen}{\{character\}}, who is \textcolor{darkgreen}{\{description\}}.
    I will provide questions from users and responses to those questions, where the responses are created in the style of you by LLMs. Based on your knowledge and experience, you must judge and assess whether the response to the question contains hallucination (fabricated or incorrect information). To evaluate hallucination, consider the following aspects:
    
    \vspace{5pt}
    (1) Whether the events, objects, locations, or persons mentioned are consistent with your established story and background. A response that is not consistent with your lore is considered as hallucination.
    
    \vspace{5pt}
    (2) Whether the response demonstrates a deep level of knowledge about a topic or concept that does not make sense for you to have, due to factors such as the topic not existing in your time period or universe. A response may refer to a topic if the question directly asks about the topic. However, answering the question with great detail would be considered as hallucination.
    
    \vspace{8pt}
    Based on the given instructions, provide a brief analysis of the given response. Then rate the response using a single score from 1 to 10, where a score of 1 represents a response with no hallucination and a 10 represents a response that does not make sense at all.

    \vspace{8pt}
    Provide your output in the following format:
    
    Analysis: <analysis>
    
    Rating: <rating>
    
    \vspace{8pt}
    \#\#\#\#\#\# Test Begin \#\#\#\#\#\#
    
    \textbf{User Question:} \textcolor{darkgreen}{\{question\}}
    
    \textbf{Response:} \textcolor{darkgreen}{\{response\}}
    
    \textbf{Output:}
    
    \end{tcolorbox}
    \caption{The prompt used for generating knowledge hallucination scores.}
    \label{fig:hallucination_prompt}
\end{figure*}

\begin{figure*}
    \centering
    \begin{tcolorbox}[
      colback=white, colframe=black, arc=3mm, width=1.1\linewidth,
      title=\textbf{Prompt for Generating Unknown Question Rejection Scores}, 
      coltitle=white, colbacktitle=gray, fonttitle=\bfseries
    ]
    Play act as \textcolor{darkgreen}{\{character\}}, who is \textcolor{darkgreen}{\{description\}}.
    I will provide questions from users and responses to those questions, where the responses are created in the style of you by other LLMs. You must judge and assess the ability whether the responses appropriately adhere to cognitive boundaries and reject questions that fall outside the scope of your knowledge and experience. 
    Use the following guidelines to assess each response:

    \vspace{3pt}
    (1)Lore consistency: If the events, objects, locations, or persons mentioned are consistent with your established story and background, the response must align with your lore.

    \vspace{3pt}
    (2)Logical Rejection: For questions about topics or concepts that does not make sense for you to have, due to factors such as topics not existing in your time period or universe, the response should logically and explicitly reject the question rather than providing detailed or fabricated
    answers.

    \vspace{5pt}
    Based on the given instructions, provide a brief analysis of the given response. Then rate the response with a binary score [0, 1], where 1 indicates that the response follows the guidelines above, 0 means not follow the guideline.

    \vspace{5pt}
    Please provide your output in the following format:
    
    Analysis: <analysis>
    
    Rating: <rating>
    
    \vspace{5pt}
    \#\#\#\#\#\# Test Begin \#\#\#\#\#\#
    
    \textbf{User Question:} \textcolor{darkgreen}{\{question\}}  
    
    \textbf{Response:} \textcolor{darkgreen}{\{response\}}    
    
    \textbf{Output:}
    \end{tcolorbox}
    \caption{The prompt used for generating unknown question rejection scores.}
    \label{fig:rejection_prompt}
\end{figure*} 

\begin{figure*}
    \centering
   \begin{tcolorbox}[
      colback=white, colframe=black, arc=3mm, width=1.1\linewidth,
      title=\textbf{Prompt for Response Generation on the Harry Potter Dataset}, 
      coltitle=white, colbacktitle=gray, fonttitle=\bfseries
    ]
    Please play as \textcolor{darkgreen}{\{character\}} in ``Harry Potter'' series and generate a response based on the dialogue context, using the tone, manner and vocabulary of \textcolor{darkgreen}{\{character\}}.
    You need to consider the following aspects to generate the character’s response:
    
    \vspace{2pt}
    (1) Feature consistency: Feature consistency emphasizes that the character always follows the preset attributes and behaviors of the character and maintains consistent identities, viewpoints, language style, personality, and others in responses.
    
    \vspace{2pt}
    (2) Character human-likeness: Characters naturally show human-like traits in dialogue, for example, using colloquial language structures, expressing emotions and desires naturally, etc.
    
    \vspace{2pt}
    (3) Response interestingness: Response interestingness focuses on engaging and creative responses. This emphasizes that the character’s responses not only provide accurate and relevant information but also incorporate humor, wit, or novelty into the expression, making the conversation not only an exchange of information but also comfort and fun.
    
    \vspace{2pt}
    (4) Dialogue fluency: Dialogue fluency measures the fluency and coherence of responses with the context. A fluent conversation is natural, coherent, and rhythmic. This means that responses should be closely related to the context of the conversation and use appropriate grammar, diction, and expressions.

    \vspace{8pt}
    Please answer in ENGLISH and keep your response simple and straightforward. If the question is beyond your knowledge, you should decline to answer and provide an explanation. Format each dialogue as: character name\textcolor{darkgreen}{\{tuple\_delimiter\}}response. Remember do not provide any content beyond the character response.
    
    \vspace{8pt}
    \#\#\#\#\#\#\#\#\#\#\#context\#\#\#\#\#\#\#\#\#\#\#\#\#\#
    
    \textcolor{darkgreen}{\{context\_data\}}
    
    \vspace{8pt}
    {-}{-}{-}{-}{-}{-}{-} Test Data {-}{-}{-}{-}{-}{-}{-}{-}{-}
    
    \textbf{Character name:} \textcolor{darkgreen}{\{character\}}
    
    \textbf{Question:} \textcolor{darkgreen}{\{question\}}
    
    \textbf{Output:}
    
    \end{tcolorbox}
    \caption{The prompt used for generating responses on the Harry Potter dataset. We use the colon character (":") for \{tuple\_delimiter\}.}
    \label{fig:generation_hp_prompt}
\end{figure*}

\end{document}